\documentclass[runningheads]{llncs}

\usepackage{eccv}

\usepackage{eccvabbrv}

\usepackage{graphicx}
\usepackage{booktabs}
\usepackage{xcolor} 

\usepackage[accsupp]{axessibility}  

\usepackage[abs]{overpic}
\usepackage{multirow}
\usepackage{array}
\usepackage{amsmath}
\usepackage{amssymb}
\usepackage[misc]{ifsym}
\usepackage[american]{babel}
\usepackage{xcolor}
\usepackage{bm}

\usepackage[pagebackref,breaklinks,colorlinks]{hyperref}

\usepackage{orcidlink}

\begin{document}

\title{SmartControl: Enhancing ControlNet for Handling Rough Visual Conditions
}

\titlerunning{SmartControl}

\author{Xiaoyu Liu\inst{1} \and
Yuxiang Wei\inst{1} \and
Ming Liu\inst{1}  \and 
Xianhui Lin \and 
Peiran Ren \and  \\
Xuansong Xie \and
Wangmeng Zuo\inst{1} }

\authorrunning{X.~Liu et al.}

\institute{$^{1}$Harbin Institute of Technology, Harbin, China}

\maketitle

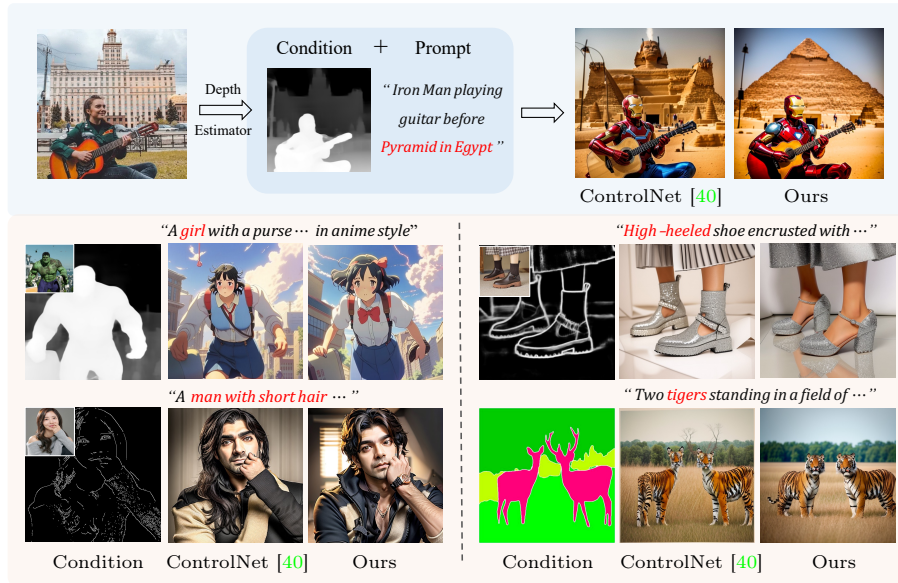
\begin{figure}[h]
\centering
 \begin{overpic}[percent, width=0.99\linewidth]{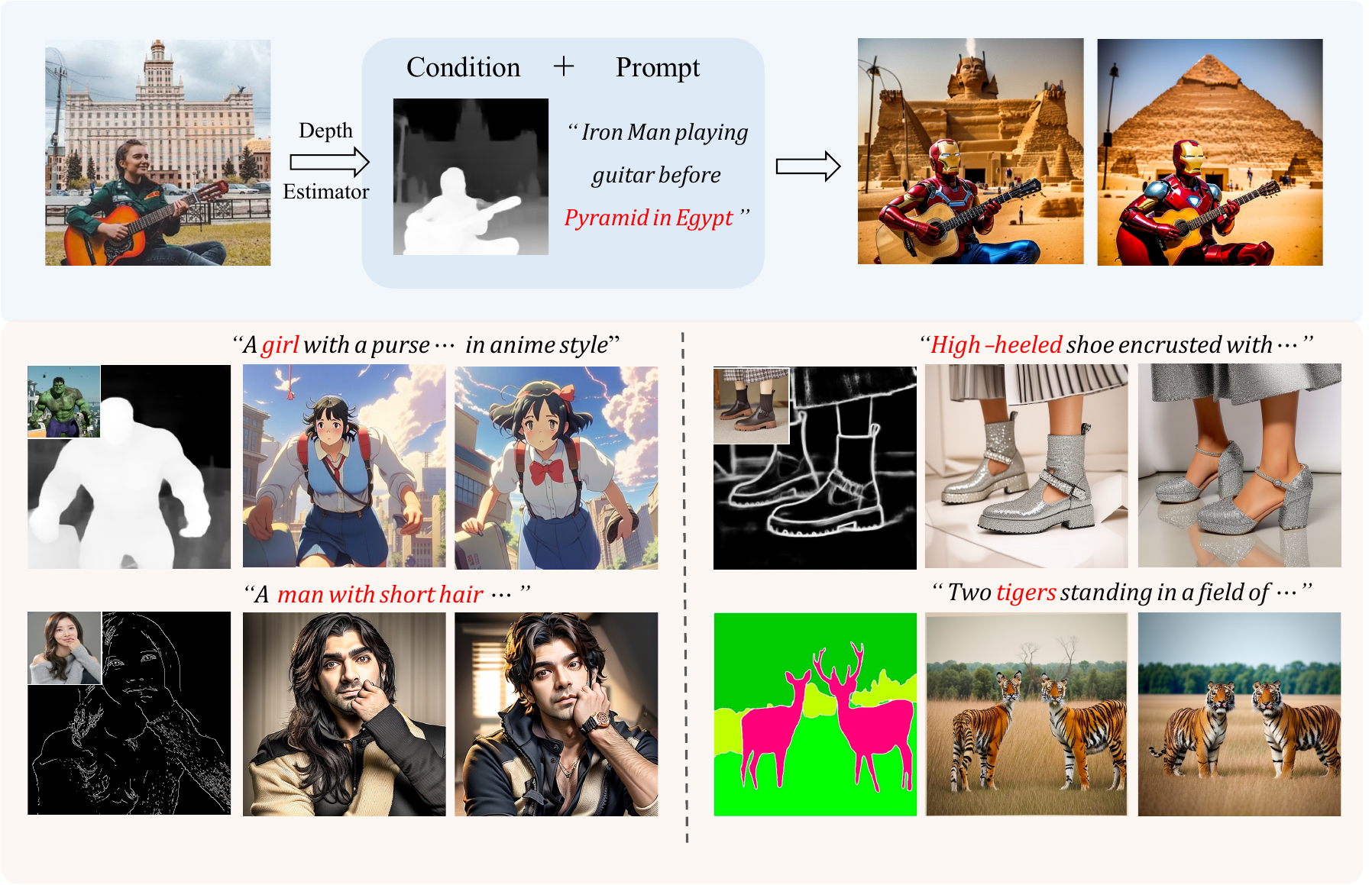}
    \put(63,42.5){\scriptsize{ControlNet~\cite{zhang2023adding}}}
    \put(85.5,42.5){\scriptsize{Ours}}
    \put(5,2.2){\scriptsize{Condition}}
    \put(17.5,2.2){\scriptsize{ControlNet~\cite{zhang2023adding}}}
    \put(38,2.2){\scriptsize{Ours}}
     \put(54.5,2.2){\scriptsize{Condition}}
    \put(67.5,2.2){\scriptsize{ControlNet~\cite{zhang2023adding}}}
    \put(88.6,2.2){\scriptsize{Ours}}
\end{overpic}
    \caption{Our proposed SmartControl can perform controllable image generation under rough visual conditions extracted from other images. In contrast, ControlNet~\cite{zhang2023adding} adheres to control conditions, which may goes against with human intentions.}
    \label{fig:SmartControl_introdution}
    \vspace{-2mm}
\end{figure}

\begin{abstract}

Human visual imagination usually begins with analogies or rough sketches.
For example, given an image with a girl playing guitar before a building, one may analogously imagine how it seems like if Iron Man playing guitar before Pyramid in Egypt.
Nonetheless, visual condition may not be precisely aligned with the imaginary result indicated by text prompt, and existing layout-controllable text-to-image (T2I) generation models is prone to producing degraded generated results with obvious artifacts.
To address this issue, we present a novel T2I generation method dubbed SmartControl, which is designed to modify the rough visual conditions for adapting to text prompt.
The key idea of our SmartControl is to relax the visual condition on the areas that are conflicted with text prompts.
In specific, a Control Scale Predictor (CSP) is designed to identify the conflict regions and predict the local control scales, while a dataset with text prompts and rough visual conditions is constructed for training CSP.
It is worth noting that, even with a limited number (e.g., 1,000$\sim$2,000) of training samples, our SmartControl can generalize well to unseen objects.
Extensive experiments on four typical visual condition types clearly show the efficacy of our SmartControl against state-of-the-arts.
Source code, pre-trained models, and datasets are available at \url{https://github.com/liuxiaoyu1104/SmartControl}.


\end{abstract}

\section{Introduction}
\label{sec:intro}

People often encounter moments of visual inspiration that ignite a desire to create compelling images by drawing on the scenes we observe.
For example, upon seeing a photograph of a girl playing guitar before a building, our imagination might spawn a novel scene, such as the modern superhero Iron Man playing the guitar set against the mysterious backdrop of the Pyramids in Egypt.
Recent layout-controllable text-to-image generation methods such as ControlNet~\cite{zhang2023adding} and T2I-Adapter~\cite{mou2023t2i}, makes it feasible to bring our imagination to life.
In practice, users first extract visual conditions (\eg, edge maps, segmentation maps, or depth) from analogously images, and combine them with carefully crafted prompts into the layout-controllable models to generate the desired mental imagery.
Nevertheless, the conditions derived from existing images or sketches may not align precisely with our mental pictures, such as the architectural discrepancies between the building and the Pyramids.
Existing layout-controllable models are trained to generate images strictly follow the visual conditions, and is prone to produce degraded  results with obvious artifacts (see the top of Fig.~\ref{fig:SmartControl_introdution}).

To improve the quality of generation on the rough condition, one possible solution is to relax the restriction of visual condition.
For example, LooseControl~\cite{bhat2023loosecontrol} proposes to control the layout of the image through a 3D bounding box, including the position, orientation, and size of the object.
Although LooseControl achieves flexible controllability, its visual condition is too loose to control the posture and actions of the objects effectively.
Another possible way is to reduce the control intensity of visual conditions.
In ControlNet~\cite{zhang2023adding}, the visual conditions are integrated into the generation process by adding the feature of visual conditions to the latent image features.
Therefore, we can decrease the fusion weight of visual conditions to relax its influence, so that the generative models can balance information from text and visual conditions.
As shown in \cref{fig:SmartControl_scale}, a proper weight may produce a desired result.
%
However, the optimal weights for different inputs are varied, and one should manually navigate all control intensities for selecting a suitable one.
Moreover, it is even infeasible to find a suitable weight for some cases (see the third row of \cref{fig:SmartControl_scale}).
Furthermore, the fusion weight is a global parameter that affects the entire image, leading to compromises between different local areas.

In this work, we propose SmartControl, an automated and flexible method for photo-realistic image generation with the text prompt and a rough visual condition as inputs.
We argue that text prompts more faithfully reflect user intentions, while visual conditions usually provide only coarse layout information.
Therefore, the key idea of our SmartControl is to relax the constraints on areas that conflict with the text prompts in the rough visual conditions.
Specifically, we propose a Control Scale Predictor (CSP) to identify the conflict regions and predict the local control scale map based on the visual conditions and text prompts.
Furthermore, to enhance the comprehension of the conflict between visual conditions and text prompts, we extract relevant priors from ControlNet~\cite{zhang2023adding} as the input of our CSP.
Finally, the predicted control scale map is employed to adaptively integrate control information into the generation process, thereby crafting our mental images.
For training the control scale predictor, a dataset with text prompts and rough visual conditions is constructed.
Thanks to the generative prior extracted from the pre-trained generative models, a limited number (\eg, 1,000$\sim$2,000) of samples is sufficient, and our SmartControl shows preferable generalization abilities to unseen objects.
%
As shown in \cref{fig:SmartControl_introdution}, our SmartControl could generate photo-realistic images faithful to text prompts while preserving useful information from the rough visual conditions.

Extensive experiments are conducted on various backbone generative models and visual condition types, and our SmartControl can perform favorably against state-of-the-art methods. 
Our contributions are listed as follows:
\begin{itemize}
    \item We present an automated and flexible text-to-image generation method under rough visual conditions (dubbed SmartControl), which achieves local-adaptive control intensities based on the inconsistency between text prompts and the visual conditions.
    \item A control scale predictor is designed to distinguish and identify conflicts between text prompts and visual conditions.
    \item A dataset with text prompts and unaligned rough visual conditions is constructed, based on which extensive experiments are conducted, showing that our proposed method performs favorably against state-of-the-art methods.
\end{itemize}

\section{Related Work}

\subsection{Text-to-Image Diffusion Model}
Diffusion models~\cite{ho2020denoising, song2020score} have achieved remarkable success in the field of text-to-image (T2I) generation~\cite{nichol2021glide,ramesh2022hierarchical,rombach2022high, saharia2022photorealistic, balaji2022ediffi}, capable of generating images with high fidelity and diversity.
T2I diffusion models redefine the image generation task as an iterative denoising process guided on text embeddings produced by language encoders such as CLIP~\cite{radford2021learning} or T5-pretrained~\cite{raffel2020exploring}.
Some methods~\cite{ramesh2022hierarchical,balaji2022ediffi,saharia2022photorealistic} adopt low-resolution models in pixel
domain, coupled with cascaded super-resolution diffusion models.
On the other hand, latent diffusion models~\cite{rombach2022high, xue2023raphael} focus exclusively on performing diffusion processes in the latent space, relying on separately trained high-resolution autoencoders.
Stable Diffusion~\cite{stablediffusion} represents a large-scale implementation of the latent diffusion model, which has been widely adopted in various applications, such as controllable image generation~\cite{mou2023t2i, zhang2023adding,huang2023composer}, customized image generation~\cite{gal2022image,ruiz2023dreambooth,wei2023elite}, image manipulation~\cite{hertz2022prompt,brooks2023instructpix2pix,parmar2023zero,mokady2023null}, and video generation~\cite{guo2023animatediff,zhang2023controlvideo,wang2023lavie,zhang2024videoelevator}.

\subsection{Controllable Text-to-Image Generation}
Text-to-image diffusion models have achieved promising ability in generating high-fidelity images based on text prompts.
However, conveying the desired spatial information solely through text prompts remains a significant challenge.
To address this, several approaches have been developed to achieve controllable text-to-image generation by adding conditional control such as pose~\cite{ju2023humansd, bhunia2023person}, 2D bounding boxes~\cite{phung2023grounded}, segmentation map~\cite{kim2023dense, xue2023freestyle,avrahami2023spatext}, and multiple conditions~\cite{mou2023t2i, zhang2023adding,huang2023composer} like edge maps, depth maps, segmentation masks, normal maps, and OpenPose.

ControlNet~\cite{zhang2023adding} adds visual conditions
to a pretrained text-to-image diffusion model through the fine-tuning of trainable encoder copies.
T2I-Adapter~\cite{mou2023t2i} employs various adapters under different conditions to achieve controllable guidance.
Several works have built upon ControlNet~\cite{zhang2023adding} to introduce improvements, including mixing modalities~\cite{hu2023cocktail, qin2023unicontrol}, efficient architecture~\cite{zavadski2023controlnet}, and loose control~\cite{bhat2023loosecontrol}.
Cocktail~\cite{hu2023cocktail} allows for the combination of existing modalities and automatically balances the differences between them.
UniControl~\cite{qin2023unicontrol} employs a mixture of expert style adapter and a task-aware HyperNet to unify various Condition-to-Image tasks in a single framework, thus compressing the model size.
ControlNet-XS~\cite{zavadski2023controlnet} focuses on designing an efficient and effective architecture without information transmission delays.
LooseControl~\cite{bhat2023loosecontrol} presents a novel approach to guiding image generation using 3D box depth conditions, employing generalized guidance to enhance the creative possibilities available to users.
However, this approach is overly permissive, focusing only on maintaining position and size, while often neglecting the crucial aspect of pose.
Unlike previous methods, FreeControl~\cite{mo2023freecontrol} provides a training-free approach for multi-condition T2I generation, enabling structural alignment with guidance images and appearance alignment with images generated without control.
In comparison to the aforementioned controllable T2I method, our solution has the ability to handle rough conditions, thereby ensuring greater flexibility in image generation. 

\section{Preliminary}

The existing layout-controllable T2I generation methods provide opportunities for human visual imagination.
However, in practice, preparing a visual condition that precisely aligns with the text prompt and user intentions is difficult or infeasible for ordinary users.
Therefore, the visual condition $\mathbf{c}$ is often obtained via cheaper ways, \eg,  extracting from an existing analogous image or inexact sketches.
We refer to such $\mathbf{c}$ as rough visual conditions (denoted by $\mathbf{c}_\mathit{rough}$) since they are not precisely aligned with the text prompt and the users usually intend to follow these conditions at a coarse scale.
To achieve the flexible generation under rough conditions, we provide a brief introduction to preliminary knowledge and exploration about the existing controllable T2I generation methods.

\subsection{ControlNet}

Stable Diffusion~\cite{stablediffusion} is a widely employed text-to-image generation method.
Given a text prompt $\mathbf{p}$, Stable Diffusion~\cite{stablediffusion} gradually integrates $\mathbf{p}$ into the image generation process via a text-conditioned cross-attention mechanism.
For controlling the layout of the generated images, alongside the pre-trained Stable Diffusion~\cite{stablediffusion}, ControlNet~\cite{zhang2023adding} further introduces a visual condition $\mathbf{c}$, which can be in the form of edge maps, segmentation masks, and so on.
Then the image generation process can be defined by $\mathbf{I} = \mathit{G}(\mathbf{p}, \mathbf{c})$, and the working scheme of ControlNet~\cite{zhang2023adding} at layer $\mathit{i}$ of the decoder $\mathit{D}$ can be represented by,
\begin{equation}
    \mathbf{h}^{\mathit{i}+1}=\mathit{D}^\mathit{i}(\mathbf{h}^\mathit{i} + \mathbf{h}^\mathit{i}_\mathit{cond}),\quad 0 \leq \mathit{i} \leq \mathit{N} - 1,
    \label{eqn:SmartControl_ControlNet_layer_op}
\end{equation}
where $\mathbf{h}^\mathit{i}$ is the feature in the $\mathit{i}$-th layer of the Stable Diffusion decoder, and $\mathbf{h}^\mathit{i}_\mathit{cond}$ is the feature generated from the visual condition $\mathbf{c}$.
In this way, the visual condition is successfully introduced into the generation process, and the images generated by ControlNet~\cite{zhang2023adding} are constrained to follow both $\mathbf{p}$ and $\mathbf{c}$.

\subsection{Control Scale Exploration}
\label{sec:SmartControl_controlscale_exploration}

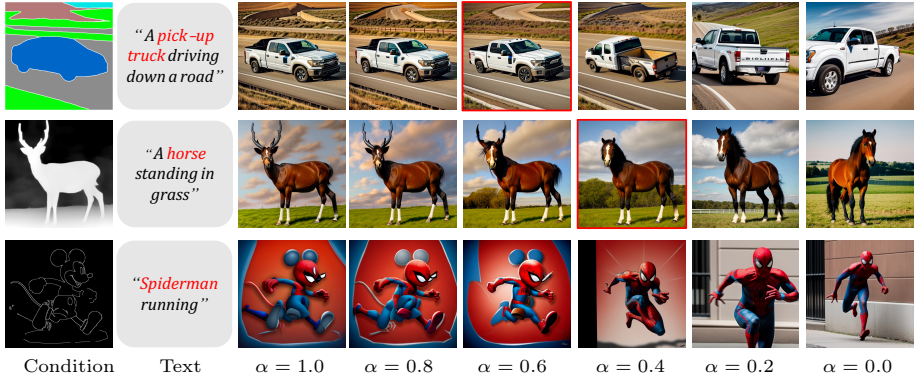
\begin{figure}[!t]
\centering
 \begin{overpic}[percent,width=0.99\linewidth]{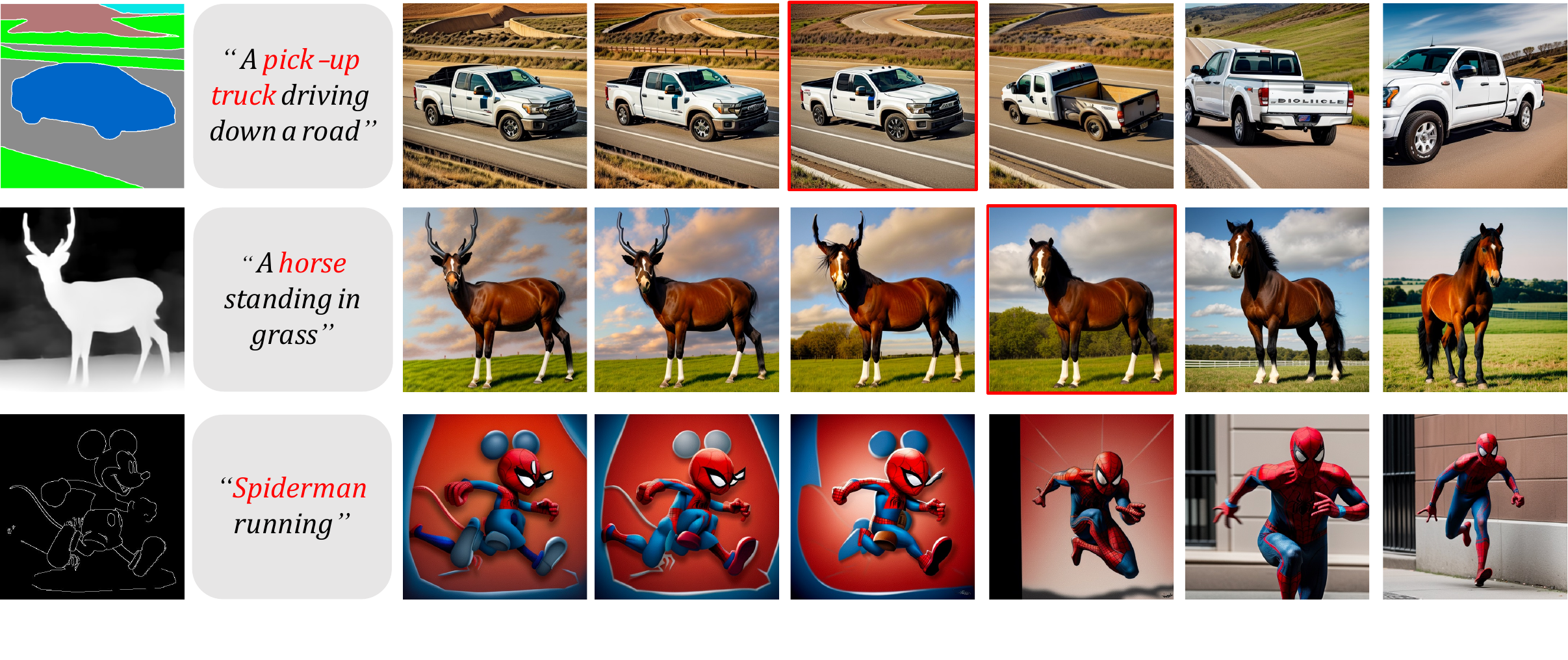}
\put(2,1.2){\scriptsize{Condition}}
\put(17,1.2){\scriptsize{Text}}
\put(27.5,1.2){\scriptsize{$\alpha=1.0$}}
\put(39.5,1.2){\scriptsize{$\alpha=0.8$}}
\put(52,1.2){\scriptsize{$\alpha=0.6$}}
\put(65,1.2){\scriptsize{$\alpha=0.4$}}
\put(77,1.2){\scriptsize{$\alpha=0.2$}}
\put(90,1.2){\scriptsize{$\alpha=0.0$}}

\end{overpic}
    \caption{ Images generated with different control scale. The plausible images are highlighted in red boxes with different control scale. And it is even infeasible to find suitable control scale for some cases.}
    \label{fig:SmartControl_scale}
    \vspace{-1mm}
\end{figure}

In ControlNet~\cite{zhang2023adding}, we can tune the fusion weight of visual conditions (\ie, $\mathbf{h} + \alpha * \mathbf{h}_\mathit{cond}$) to relax the inconsistency between the text prompts and the visual conditions.
We vary the value of $\alpha$ from 1.0 to 0.0 to explorate its influence.

As shown in \cref{fig:SmartControl_scale}, when given a pair of an unaligned text prompt $\mathbf{p}$ and a rough visual condition $\mathbf{c}_\textit{rough}$, ControlNet~\cite{zhang2023adding} (\ie, when $\alpha=1.0$) strictly follows the layout of $\mathbf{c}_\mathit{rough}$ and fits the object mentioned in the text prompt $\mathbf{p}$ into the shape described by $\mathbf{c}_\mathit{rough}$.
For example, a deer antler is added to the horse, and two round ears appear on the head of Spider-Man.
By gradually decreasing the value of $\alpha$, one can see that the generated images become better aligned with the text prompt $\mathbf{p}$, until the effect of $\mathbf{c}_\mathit{rough}$ disappears when $\alpha=0.0$.

Besides, we have also observed a large amount of samples, drawing the following conclusions.
(i)~For a portion of the $(\mathbf{p}, \mathbf{c}_\mathit{rough})$ pairs, a proper control scale $\alpha$ can be found to generate a plausible image\footnote{For the $(\mathbf{p}, \mathbf{c}_\mathit{rough})$ pairs we delicately prepared in \cref{sec:SmartControl_dataset}, we can find a suitable $\alpha$ for around 60\%$\sim$70\% of the samples.}, but it varies for different visual conditions and text prompts.
(ii)~Even if the optimal global $\alpha$ is not found, it seems promising to obtain a desired image by combining results with different $\alpha$.
For example, as shown in the third row of \cref{fig:SmartControl_scale}, we can get a potential result by combining Spider-Man when $\alpha=0.6$ and background when $\alpha=0.4$.
(iii)~For areas that $\mathbf{c}_\mathit{rough}$ conflicts with $\mathbf{p}$, a large enough freedom (\ie, small enough $\alpha$) should be assigned to breaking free from the constraints of $\mathbf{c}_\mathit{rough}$. On the contrary, a sufficiently large $\alpha$ should be set in other areas to ensure the effectiveness of $\mathbf{c}_\mathit{rough}$.

\section{Proposed Method}

In our work, we propose SmartControl which is designed to modify the rough visual conditions for adapting to text prompt.
Specifically, we design a control scale predictor $\mathit{f}$ to predict spatial adaptive control scale map $\bm{\alpha}$ (referenced in \cref{sec:SmartControl_predictor}).
The predicted scale map is employed to adaptively integrate control information into the generation process for generating the plausible images.
To train such a control scale predictor, we will construct an unaligned text-condition training dataset (as detailed in \cref{sec:SmartControl_dataset}).
Finally, we will introduce the learning objective utilized to train our SmartControl in \cref{sec:SmartControl_loss}.

\begin{figure}[!t]
\centering
 \begin{overpic}[width=0.99\linewidth]{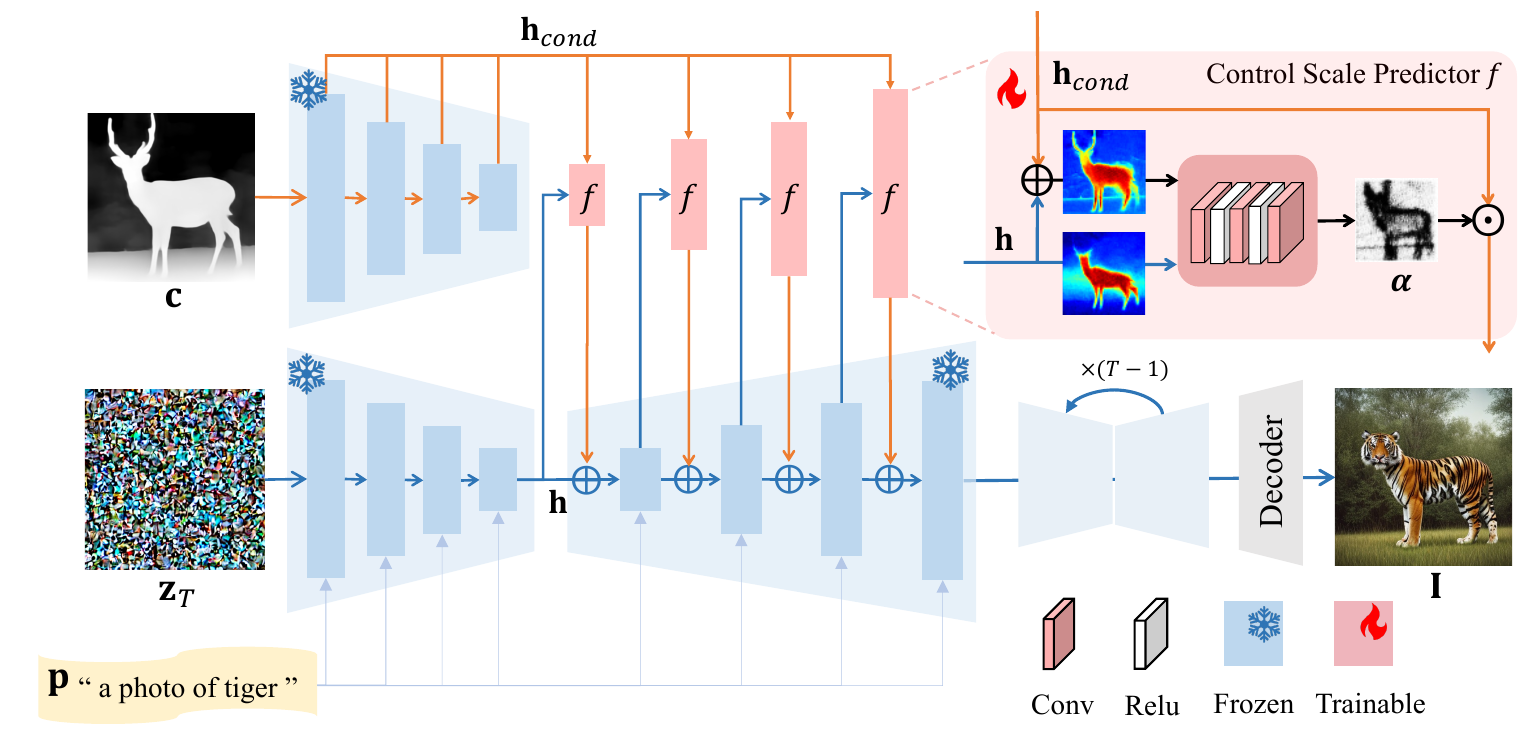}

\end{overpic}
    \caption{Framework of proposed SmartControl. 
    Our method is built upon ControlNet, and can generate photo-realistic images with inconsistent prompt and rough visual condition (\ie, tiger \textit{v.s.} deer) as input.
    To achieve this, we introduce a control scale predictor $\mathit{f}$ for each decoder block of ControlNet. 
    The predictor takes $\mathbf{h}$ and $\mathbf{h}+ \mathbf{h}_\mathit{cond}$ as input and predicts a pixel-wise control scale map $\boldsymbol{\alpha}$. 
    The condition feature $\mathbf{h}_\mathit{cond}$ is then updated by $\bm{\alpha}\cdot\mathbf{h}_\mathit{cond}$ to relax the control scale at conflict region, resulting a plausible and photo-realistic generated image.
    }
    \label{fig:SmartControl_pipline}
    \vspace{-1mm}
\end{figure}

\subsection{Control Scale Predictor}
\label{sec:SmartControl_predictor}
 Based on the observations in \cref{sec:SmartControl_controlscale_exploration}, it is evident that manually filtering the control scale is challenging, and these global scale is insufficient to achieve the precise control of different regions.
These observations inspire us to introduce a control scale predictor to predict local control scale map without the need for manual selection. 
In this section, we will introduce the details of the predictor, including input representation and network architecture.

\noindent\textbf{Input Representation.}
Based on the third observations in \cref{sec:SmartControl_controlscale_exploration}, the extent that $\mathbf{c}_\mathit{rough}$ and $\mathbf{p}$ conflict determines the value of $\bm{\alpha}$.
Therefore, our predictor need to comprehend and align the text prompts and conditions for predicting scale map $\bm{\alpha}=\mathit{f}(\mathbf{p}, \mathbf{c}_\mathit{rough}; \theta)$.
%
Such the ability to handle two different modalities, which requires a substantial dataset for training.
To reduce the training requirements, we propose leveraging the superior capabilities of ControlNet as a prior. Building upon Stable Diffusion~\cite{rombach2022high}, ControlNet can extract meaningful features from prompts and visual conditions separately and utilize them to generate desired images.
Specifically, $\mathbf{h} + \mathbf{h}_\mathit{cond}$ integrates information about the visual condition, while $\mathbf{h}$ encodes the information from the given prompt.
Therefore, instead of using $\mathbf{p}$ and $\mathbf{c}$ as inputs to the predictor, we utilize $\mathbf{h}$ and $\mathbf{h} + \mathbf{h}_\mathit{cond}$, which facilitates the easier identification of inconsistencies.

\noindent\textbf{Network Architecture.}
The overall architecture of proposed SmartControl is illustrated in \cref{fig:SmartControl_pipline}.
Within each decoder block $\mathit{D}^\mathit{i}$, we incorporate a control scale predictor $\mathit{f}^\mathit{i}$ to predict spatially adaptive control scales $\bm{\alpha}^\mathit{i}$.
The control scale predictor consists of three stacked modules (each containing a convolutional layer and a ReLU layer) and a sigmoid function.
The $\mathit{i}$-th predictor takes  $\mathbf{h}^\mathit{i}$ and $\mathbf{h}^\mathit{i}+ \mathbf{h}_\mathit{cond}^\mathit{i}$ as input and predicts a pixel-wise control scale map $\boldsymbol{\alpha}^\mathit{i}$,
\begin{equation}
    \boldsymbol{\alpha}^\mathit{i} = \mathit{f}^\mathit{i}(\mathbf{h}^\mathit{i}, \mathbf{h}^\mathit{i}+ \mathbf{h}_\mathit{cond}^\mathit{i}),\quad 0 \leq \mathit{i} \leq \mathit{N} - 1.
    \label{eqn:_definition}
\end{equation}
As depicted in \cref{fig:SmartControl_pipline}, the predicted $\boldsymbol{\alpha}$ exhibits minimal values in the conflict region (antlers and legs), while approaching 1.0 in other regions (background).
This indicates that the predicted $\boldsymbol{\alpha}$ is plausible, and we can utilize $\boldsymbol{\alpha}$ to generate the desired image of a tiger.

\subsection{Unaligned Data Construction Pipeline}
\label{sec:SmartControl_dataset}

\begin{figure}[!t]
\centering
 \begin{overpic}[width=0.99\linewidth]{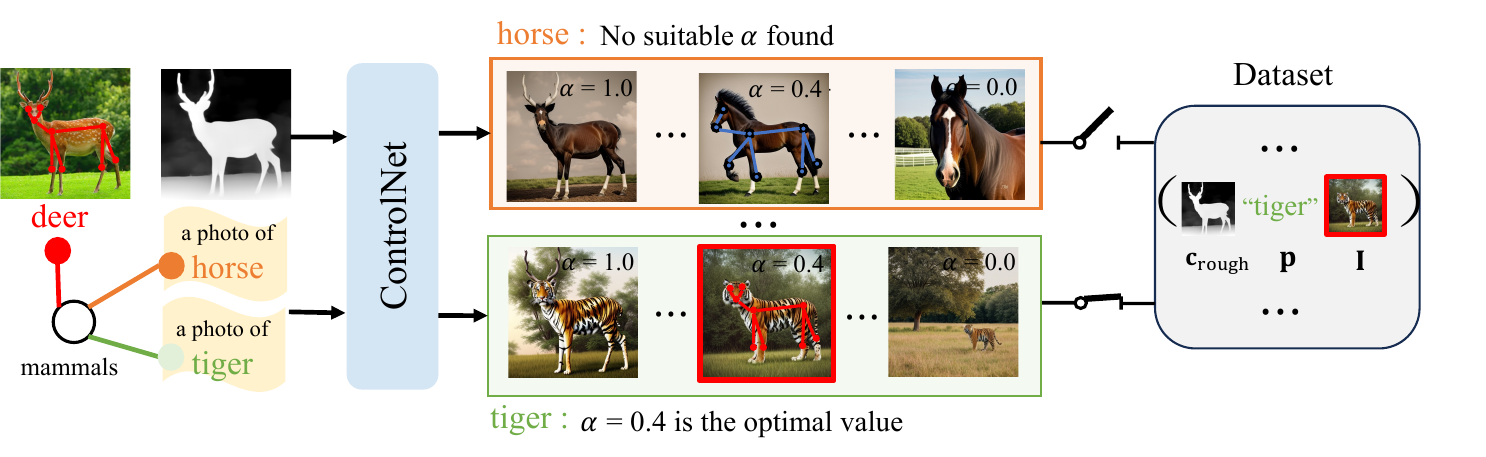}

\end{overpic}
    \caption{
    Pipeline for unaligned data construction.
    Given an image and corresponding class, we extract the visual condition $\mathbf{c}_\mathit{rough}$ (\eg, depth) by the pre-trained estimator.
    Then, for the given class (\eg, deer), we select an alternative unaligned class (\eg, tiger or horse) based on class hierarchy, and use it to obtain the unaligned prompt $\mathbf{p}$.
    By iterating through different control scale $\alpha$ of ControlNet~\cite{zhang2023adding}, we can generate a series of images for $(\mathbf{c}_\mathit{rough}, \mathbf{p})$.
    Then, we manually filter those images that are faithful to both text and rough condition to construct our dataset.
    For example, for tiger, the image generated with $\alpha = 0.4$ is plausible and is added to our dataset.
    While for horse, there is not a suitable image and all images are discarded.
    }
    \label{fig:SmartControl_dataset}
    \vspace{-1mm}
\end{figure}

 ControlNet~\cite{zhang2023adding} utilizes an image $\mathbf{I}$ as input and generates aligned conditions $\mathbf{c}$ and text prompt $\mathbf{p}$, which is not suitable to train our SmartControl.
In this section, we will introduce the workflow for constructing the unaligned text-condition dataset as shown in \cref{fig:SmartControl_dataset}.
 Specifically, we first generate the unaligned visual conditions and text.
Then, the paired image is generated by ControlNet~\cite{zhang2023adding} based on these unaligned texts and conditions.

\noindent\textbf{Generating Unaligned Visual Conditions and Text Prompts.}
The original image is from OpenImage~\cite{kuznetsova2020open} and contains an object occupying over 30\% of the image area.
The visual condition is generated by a pre-existing condition estimator.
To create a plausible inconsistent text prompt for the given image with class <$\texttt{cls}_{init}$>, we first employ the hierarchical class tree to determine one alternative class <$\texttt{cls}_{alt}$>, where <$\texttt{cls}_{init}$> and <$\texttt{cls}_{alt}$> share the same parent class.
Then, the inconsistent text prompt can be formatted as ``\texttt{a photo of a <$\text{cls}_{alt}$>}.''.
For example, as shown in \cref{fig:SmartControl_dataset}, if the rough depth condition $\mathbf{c}_\mathit{rough}$ is from an image of a deer, the corresponding prompt $\mathbf{p}$ is ``\texttt{a photo of a horse.}''.

\noindent\textbf{Generating Paired Images Based on ControlNet~\cite{zhang2023adding}.}
From the above analysis, we iterate over different control scale $\alpha$, \ie, $\alpha \in \{1.0,0.8,0.6,0.4,0.2,0.0\}$, to perform sampling based on ControlNet~\cite{zhang2023adding}, followed by manual filtering.
If a proper control scale $\alpha$ can be found to generate a plausible image, we add this data into our dataset.
In cases where no proper $\alpha$ is found, the data is discarded.
It is worth noting that our iteration is limited within a dilation range of the chosen object and the value of $\alpha$ is 1.0 in other areas.
Besides, we also acquire $\mathbf{m}_\mathit{conflict}$ and $\mathbf{m}_\mathit{bg}$ for training. 
$\mathbf{m}_\mathit{conflict}$ denotes the areas of conflict between $\mathbf{c}_\mathit{rough}$ and $\mathbf{p}$, \ie, different part between the mask of <$\text{cls}_{init}$> and <$\text{cls}_{alt}$>, while $\mathbf{m}_\mathit{bg}$ represents the background region, defined as follows,
\begin{equation}
    \mathbf{m}_\mathit{conflict}=|\mathbf{m}_\mathit{alt}-\mathbf{m}_\mathit{init}|, \mathbf{m}_\mathit{bg}=1-(\mathbf{m}_\mathit{alt} \lor \mathbf{m}_\mathit{init}),
\end{equation}
where $\mathbf{m}_\mathit{alt}$ and $\mathbf{m}_\mathit{init}$ are obtained based on the existing segmentation method~\cite{zou2024segment} based on ``\texttt{a photo of a <$\text{cls}_{alt}$>.}'' and ``\texttt{a photo of a <$\text{cls}_{init}$>.}''.

\subsection{Learning Objective.}
\label{sec:SmartControl_loss}
Following ControlNet~\cite{zhang2023adding}, we employ the mean-squared loss to train our predictor,
\begin{equation}
    \mathcal{L}_\mathrm{LDM} = \mathbb{E}_{\mathbf{z}_0, \mathbf{t}, \mathbf{p}, \mathbf{c}_\mathit{rough}, \epsilon \sim \mathcal{N}(0,1)}[\| \epsilon-\epsilon_\theta(\mathbf{z}_\mathit{t}, \mathbf{t}, \mathbf{p}, \mathbf{c}_\mathit{rough})) \|_2^2],
    \label{eqn:SmartControl_ldmloss}
\end{equation}
where $\epsilon_\theta$ denotes our model and $\mathbf{z}_0$ represents the latent embedding of real image.
$\epsilon$ denotes the unscaled noise and $\mathbf{t}$ denotes the time step of diffusion process.
$\mathbf{z}_\mathit{t}$ is the latent noise at $\mathbf{t}$ step. 

To provide explicit supervision for the control scale predictor $\mathit{f}$, we further introduce a regularization term to ensure that the values of $\bm{\alpha}$ should be maintained above $\alpha_\mathit{bg}$ in the background regions and below $\alpha_\mathit{conflict}$ in the conflict regions,
\begin{equation}
    \mathcal{L}_\mathrm{c} = \mathbf{m}_\mathit{conflict} \cdot \max (\mathbf{0}, \bm{\alpha}-\alpha_\mathit{conflict}) + \mathbf{m}_\mathit{bg} \cdot \max (\mathbf{0}, \alpha_\mathit{bg} -\bm{\alpha}),
\end{equation}
where $\alpha_\mathit{bg}$ and $\alpha_\mathit{conflict}$ are hyper-parameters.
$\mathbf{m}_\mathit{conflict}$ denotes the mask of conflict areas, and $\mathbf{m}_\mathit{bg}$ is the mask of background.

The overall learning objective for training the SmartControl is defined by,
\begin{equation}
    \mathcal{L}=\mathcal{L}_\mathrm{LDM} + \lambda_\mathrm{c}\mathcal{L}_\mathrm{c},
\end{equation}
where $\lambda_\mathrm{c}$ is hyper-parameters for balancing different loss terms.

\section{Experiments}
\label{sec:blind}

\subsection{Experimental Details}

\noindent\textbf{Datasets}.
We collect training datasets across four types of conditions including depth, HED, segmentation, and canny.
The dataset sizes for each condition are 2,000, 1,500, 1,500, and 1,000 images respectively.
For each type, we collect an evaluation dataset of 100 images including 70 images with significant conflicts, 20 images with mild conflicts, and 10 conflict-free images to assess the performance in handling diverse conditions.
Our evaluation dataset includes 48 classes, and 12.5\% of those classes do not appear in the training dataset, which allows us to evaluate the generalization ability.

\vspace{0.5mm}
\noindent\textbf{Evaluation Metrics}.
Following~\cite{tumanyan2022plugand}, we use \textit{CLIP Score}~\cite{radford2021learning} metric to measure text-image alignment and use \textit{Self-similarity distance} metric to measure the structural similarity between two images in the feature space of the DINO-ViT model~\cite{caron2021emerging}.
A smaller Self-similarity distance implies that the generated image closely preserves the structure of the source image (\ie, the image that provides rough condition).
Moreover, we introduce a metric named \textit{Class Confidence} to assess whether the generated images align with the desired class.
A higher \textit{Class Confidence} indicates that the generated images closely match the desired class, not affected by the inherent class of the rough conditions.
To comprehensively evaluate structure preservation and image-text alignment, we propose to utilize GPT-4V~\cite{achiam2023gpt} as a novel metric.
Given two images from different methods, we ask GPT-4V~\cite{achiam2023gpt} to determine which of them is better by examining through two aspects: first, whether the pose or layout matches the condition image, and second, whether it aligns more accurately with the given text.
The specific prompt can be found in the supplementary material.

\vspace{0.5mm}
\noindent\textbf{Implementation Details}.
In all our experiments, we train our control scale predictor based on the pre-trained ControlNet~\cite{zhang2023adding}, while keeping all parameters of ControlNet~\cite{zhang2023adding} fixed.
The model is trained with an AdamW~\cite{loshchilov2017decoupled} optimizer with weight decay of $1\times10^{-5}$ for 200 epochs.
The trade-off parameter $\lambda_\mathrm{c}$ is determined to be 0.01. 
Furthermore, $\alpha_\mathit{conflict}$ and $\alpha_\mathit{bg}$ are set at 0.2 and 0.8 respectively.

\subsection{Comparison with Existing Methods}

We choose the following state-of-the-art models in controllable image generation as competing methods: ControlNet~\cite{zhang2023adding}, T2I-Adapter~\cite{mou2023t2i}, and Uni-ControlNet~\cite{qin2023unicontrol}. 
However, standard ControlNet~\cite{zhang2023adding} and T2I-Adapter~\cite{mou2023t2i} are not suitable for rough conditions. 
For a fair comparison, we employ the small control scale $\alpha_{fix}$, for both ControlNet~\cite{zhang2023adding} and T2I-Adapter~\cite{mou2023t2i}. 
Here, $\alpha_{fix}$ represents the optimal but fixed control scale for the entire evaluation dataset.
However, $\alpha_{fix}$ varies across different modalities.
For example, we use $\alpha_{fix}=0.4$ for the depth condition and $\alpha_{fix}=0.6$ for the segmentation condition in ControlNet~\cite{zhang2023adding}.

\vspace{0.5mm}
\noindent\textbf{Quantitative Comparison}.
We conduct comprehensive experiments in four types of conditions to assess the effectiveness of the proposed method, and the quantitative results are shown in \cref{tab:comparisonsmart}.
We can observe that while ControlNet ($\alpha$=1.0)~\cite{zhang2023adding} and T2I-Adapter($\alpha = 1.0$)~\cite{mou2023t2i} are stronger in maintaining structure, they struggle to generate images aligned with text prompts, resulting in significantly lower CLIP Scores.
ControlNet($\alpha=\alpha_{fix}$)~\cite{zhang2023adding} and T2I-Adapter($\alpha=\alpha_{fix}$)~\cite{mou2023t2i} show inferior performance as they exhibit limitations in handling diverse text prompts and structural conditions.
Our method exhibits significant improvement in CLIP Scores compared to the previous methods, indicating improved image-text alignment. 
Note that We did not achieve a superior Self-similarity metric in the unpaired evaluation dataset.
However, a lower Self-similarity metric may indicate that the generated images overly adhere to the rough conditions, which does not always equate to better performance.
In the supplementary material, we will provide a Self-similarity metric calculated with pseudo-ground truths instead of the source images.
Considering the significant effort for utilizing GPT-4V~\cite{achiam2023gpt} as the metric, we select the commonly used condition (\ie, depth) to compare our method with ControlNet($\alpha=\alpha_{fix}$)~\cite{zhang2023adding}.
In the majority of cases, specifically 67\%, GPT-4V~\cite{achiam2023gpt} ranked our result better.

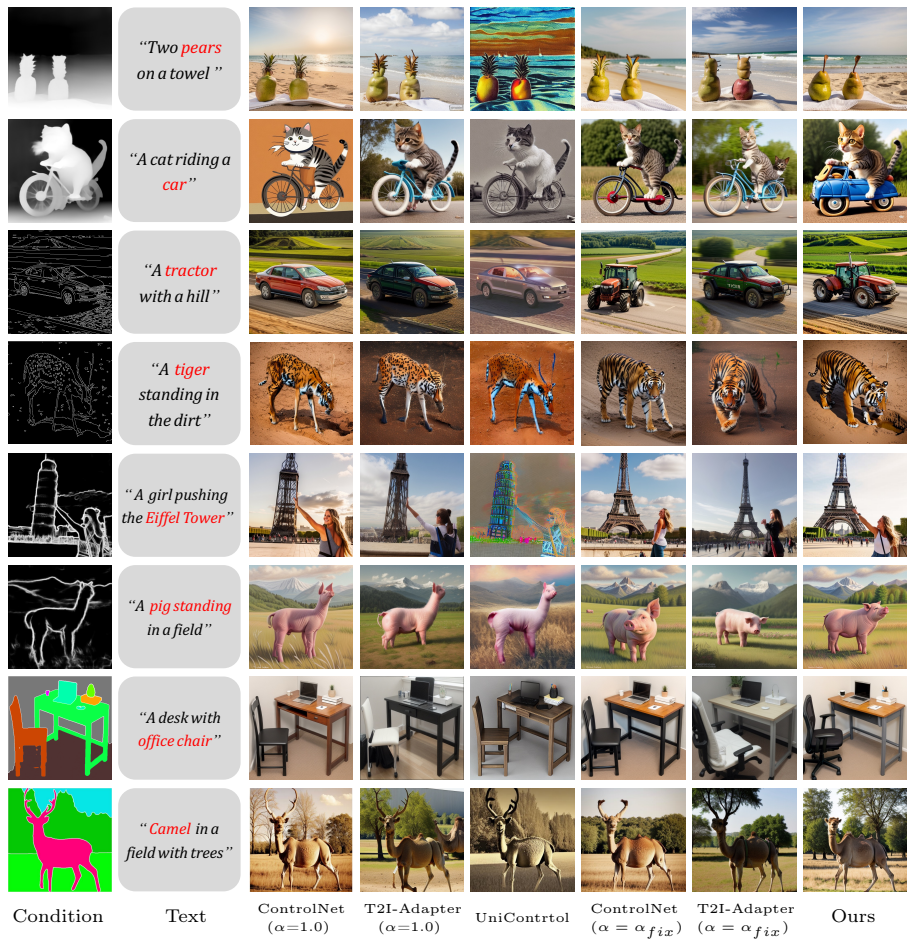
\begin{figure*}[t!]
    \centering

    \begin{overpic}[percent,width=.98\linewidth]{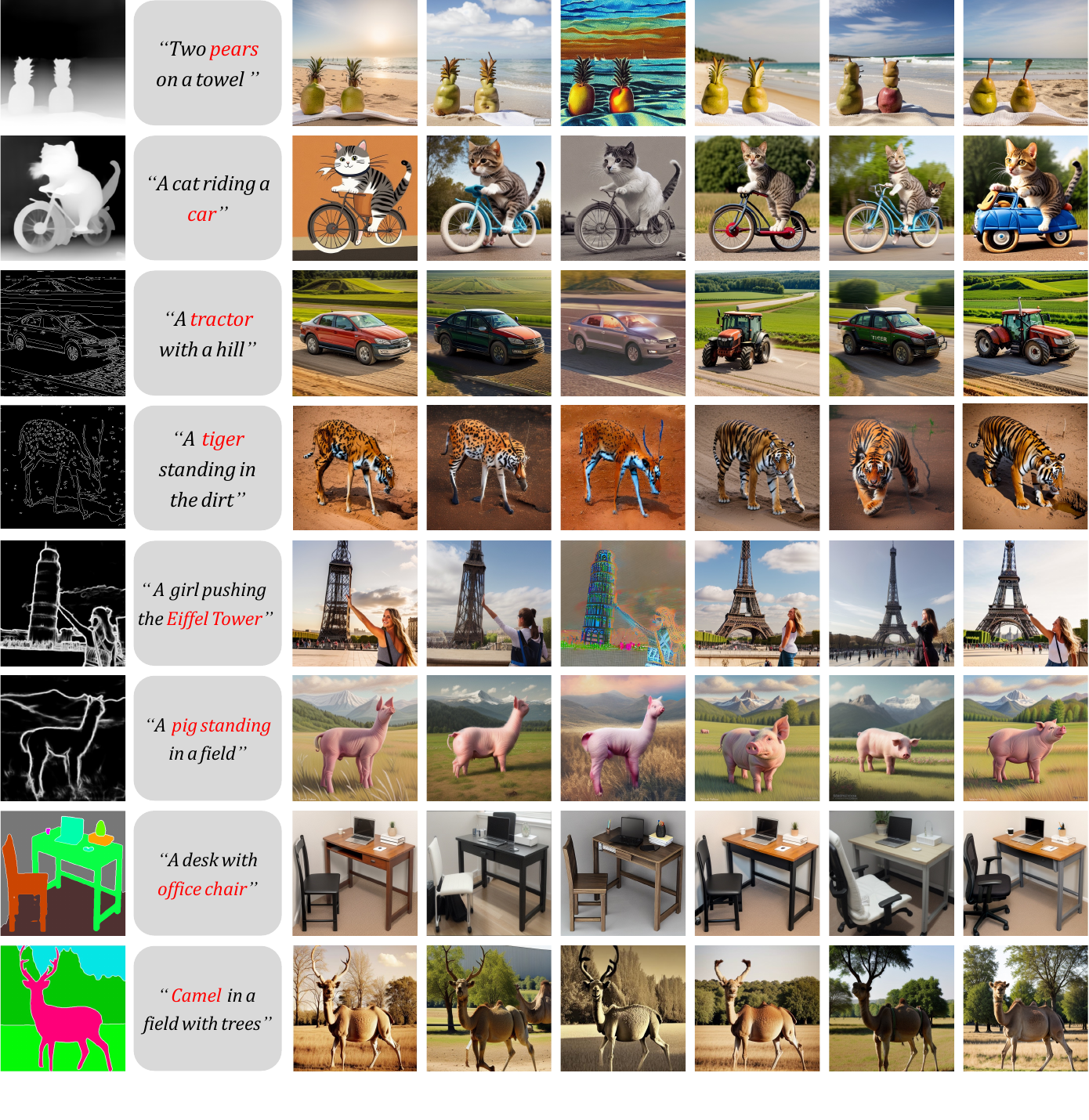}
    \put(27,2){\tiny{ControlNet}}
    \put(28,0){\tiny{($\alpha$=1.0)}}

    \put(63,2){\tiny{ControlNet}}
     \put(63,0){\tiny{($\alpha=\alpha_{fix}$)}}

     \put(38.5,2){\tiny{T2I-Adapter}}
    \put(40,0){\tiny{($\alpha$=1.0)}}
    
     \put(74.5,2){\tiny{T2I-Adapter}}
     \put(74.5,0){\tiny{($\alpha=\alpha_{fix}$)}}

     \put(50.5,1){\tiny{UniContrtol}}
     
     \put(89,1){\scriptsize{Ours}}
     \put(0.5,1){\scriptsize{Condition}}
     \put(17,1){\scriptsize{Text}}
    \end{overpic}
    \caption{Qualitative comparison with different modalities, image prompts and additional visual conditions. SmartControl achieves reasonable spatial control and superior image-text alignment compared to existing methods, resulting in a closer match to human intentions.}
    \label{fig:SmartControl_compare}
    \vspace{-1mm}
\end{figure*}

\begin{table*}[t!]
    \centering
    \caption{Quantitative comparison for Controllable Text-to-Image Generation for rough conditions on our evaluation dataset.The best results are highlighted with \textbf{bold}.}
    \scalebox{0.70}{
        \begin{tabular}{l|ccc|ccc|ccc|ccc}

        \hline
        
        \multirow{2}{*}{Method} &\multicolumn{3}{c|}{Depth} &\multicolumn{3}{c|}{Canny} &\multicolumn{3}{c|}{Seg} &\multicolumn{3}{c}{HED} \\
        
        \cline{2-13}
        & CLIP$\uparrow$  & CLASS$\uparrow$ &Sim$\downarrow$  & CLIP$\uparrow$  & CLASS$\uparrow$   &Sim$\downarrow$  & CLIP$\uparrow$  & CLASS$\uparrow$   &Sim$\downarrow$  & CLIP$\uparrow$  & CLASS$\uparrow$   &Sim$\downarrow$  \\ 
        \cline{1-13}

        ControlNet($\alpha$=1.0)~\cite{zhang2023adding}
        & 0.257& 0.602& \textbf{0.100} & 0.244& 0.467& \textbf{0.107}&
        0.258&0.666&\textbf{0.115} & 0.264& 0.647& 0.123\\
        
        T2I Adapter($\alpha$=1.0)~\cite{mou2023t2i}
        &0.267&0.593
        &0.123&0.253&0.464 & 0.109 &0.251& 0.492 & 0.138& 0.261 &0.621& 0.106 \\
    
        UniContrtol~\cite{qin2023unicontrol}
        & 0.251&0.597&0.102&0.240&0.379&0.117&0.261&0.668&0.116 & 0.227& 0.336 & \textbf{0.082} \\
        
        \hline
        ControlNet($\alpha=\alpha_{fix}$)~\cite{zhang2023adding} & 0.268 & 0.710 & 0.136 & 0.270 & \textbf{0.736} & 0.149 & 0.267 & 0.696 & 0.140 & 0.271 & 0.727 & 0.143 \\
        
        T2I Adapter($\alpha=\alpha_{fix}$)~\cite{mou2023t2i} & 0.271
        &0.721&0.137&\textbf{0.272}&0.682&0.141&0.263&0.668&0.143 & 0.269 & 0.747 & 0.137\\
           
        \hline
        Ours & \textbf{0.274}& \textbf{0.742} &0.128& \textbf{0.272}&0.721&0.143& \textbf{0.277}&\textbf{0.780}&0.140&\textbf{0.276}& \textbf{0.768} & 0.142\\
        \hline
        \end{tabular}
    }
    \label{tab:comparisonsmart}
    \vspace{-2mm}
\end{table*}

\vspace{0.5mm}
\noindent\textbf{Qualitative Comparison}.
The qualitative results of competing methods are shown in \cref{fig:SmartControl_compare}.
ControlNet ($\alpha=1.0$)~\cite{zhang2023adding}, T2I-Adapter ($\alpha=1.0$)~\cite{mou2023t2i}, and UniControl~\cite{qin2023unicontrol}, which are constrained by rough conditions, generate images that are unrealistic and misaligned with the text prompts. 
Meanwhile, ControlNet ($\alpha=\alpha_{fix}$)~\cite{zhang2023adding} and T2I-Adapter ($\alpha=\alpha_{fix}$)~\cite{mou2023t2i} offer some improvement in specific scenarios. 
However, due to the uniform $\alpha$ across all images, they still encounter failure in numerous situations.
In the example from the second row, despite altering the pose of the cat, it is not possible to successfully transform from a bicycle to a car using ControlNet ($\alpha=\alpha_{fix}$)~\cite{zhang2023adding} and T2I-Adapter ($\alpha=\alpha_{fix}$)~\cite{mou2023t2i}.
Moreover, in cases where it is necessary to remove regions (which require extremely small values of $\alpha$), such as in the eighth row example, all competing methods would result in images retaining the deer antlers.
As illustrated in \cref{fig:SmartControl_compare}, our proposed SmartControl is capable of generating images that not only closely resemble real images but also align more accurately with text prompts and useful information in rough conditions, demonstrating its superior performance.
More qualitative results are given in the supplementary material.
\begin{figure}[t]
\centering
 \begin{overpic}[percent,width=0.99\linewidth]{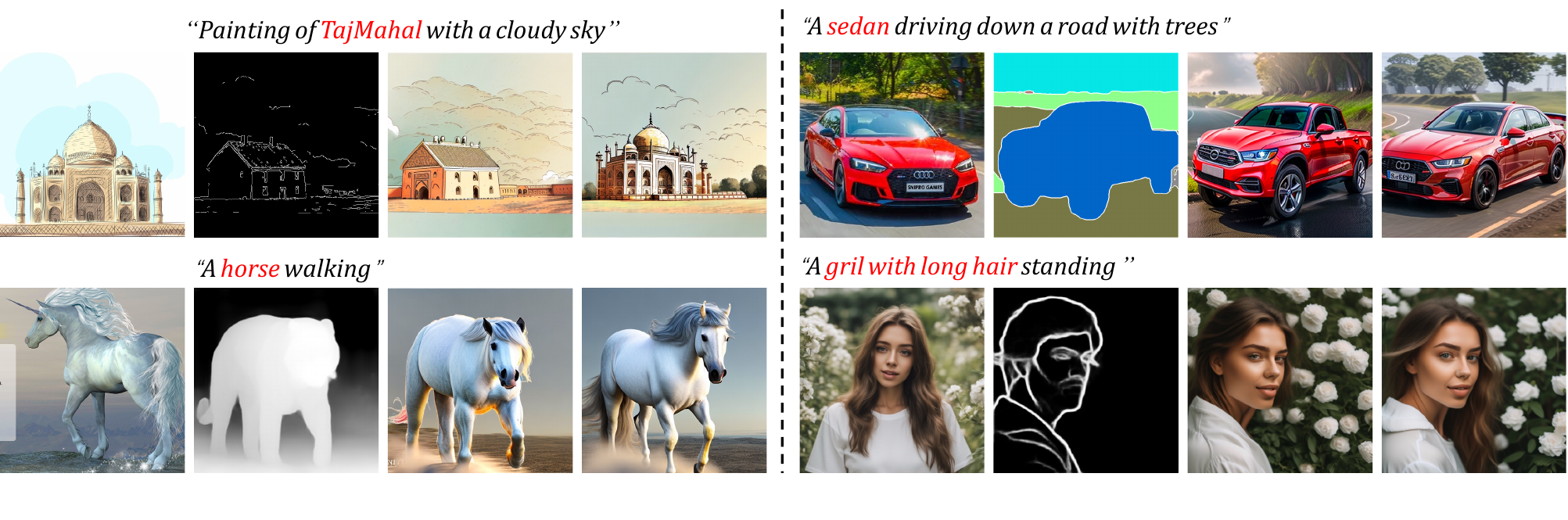}
 \put(25,1){\tiny{IP-Adapter+}}
 \put(25.5,-1){\tiny{ControlNet}}
 \put(38,1){\tiny{IP-Adapter+}}
 \put(41,-1){\tiny{Ours}}
 \put(-0.5,0){\tiny{Image Prompt}}
 \put(14,0){\tiny{Condition}}

 \put(76.5,-1){\tiny{ControlNet}}
 \put(76,1){\tiny{IP-Adapter+}}
 \put(92,-1){\tiny{Ours}}
 \put(89,1){\tiny{IP-Adapter+}}
 \put(50.5,0){\tiny{Image Prompt}}
 \put(65,0){\tiny{Condition}}
    \end{overpic}
    \caption{Visualization of generated samples with the IP-Adapterr~\cite{ye2023ip}. Note that we do not need fine-tune our control scale predictor.}
    \label{fig:SmartControl_IPAdapter}
    \vspace{-1mm}
\end{figure}
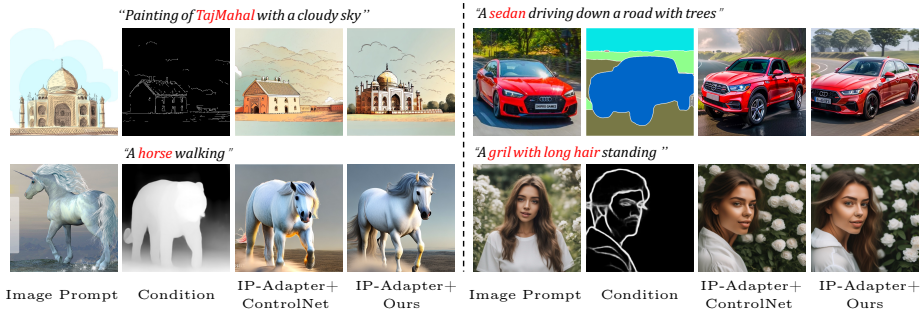
%

\vspace{0.5mm}
\noindent\textbf{User Study.}
We invite 20 users to participate in our user study to assess the effectiveness of our methods. 
We utilize six different methods and generate 40 images for each method based on different types of visual conditions and text prompts.
Each user is requested to select the best image based on the text-image alignment and structural similarity with visual conditions.
In the majority of cases, \ie, 78.3\%, users prefer our
method.

\subsection{More Results}
While originally designed for rough conditions, SmartControl demonstrates robust generalization capabilities, enabling it to effortlessly adapt to other models without retraining. 
In this section, we showcase additional results through the integration of our SmartControl with the IP-Adapter~\cite{ye2023ip}.
The primary function of IP-Adapter~\cite{ye2023ip} is to interpret image prompts to pre-trained text-to-image diffusion models. 
 \cref{fig:SmartControl_IPAdapter} shows that the images generated by SmartControl are not only more captivating but also more coherent with image prompts under rough conditions.

\subsection{Ablation Study}
\begin{figure}[t!]
    \centering
    \begin{minipage}[t]{0.49\textwidth}
    \centering
    \begin{overpic}[percent,width=.99\textwidth]{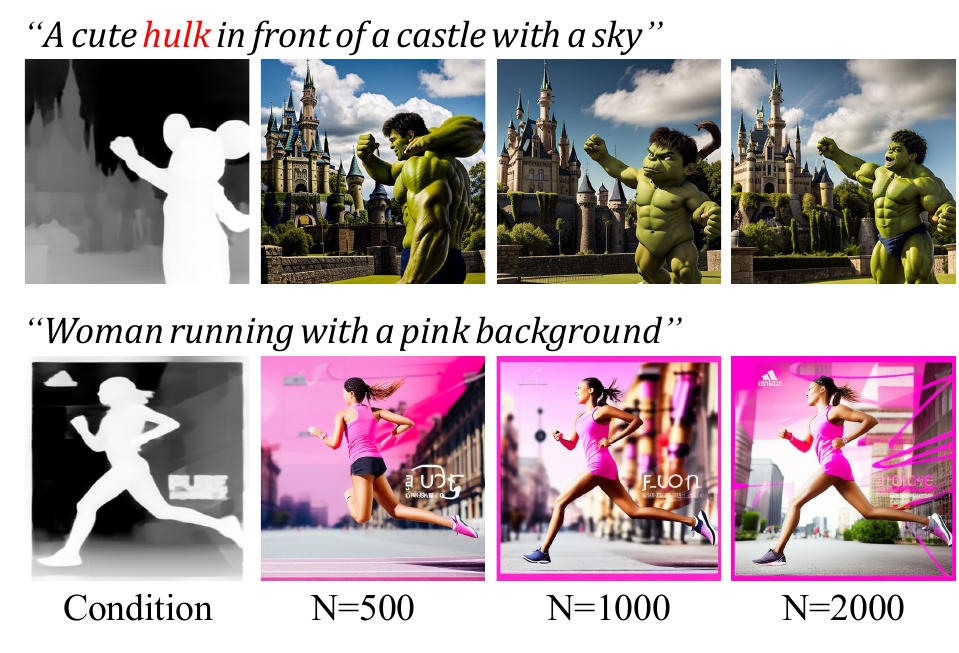}
    \end{overpic}
    \vspace{-6mm}
    \caption{\scriptsize{Visualisation of ablation study different training dataset sizes.}}
 \label{fig:SmartControl_datanum}
 \end{minipage}
 \hfill
 \begin{minipage}[t]{0.49\textwidth}
    \centering
    \includegraphics[width=\textwidth]{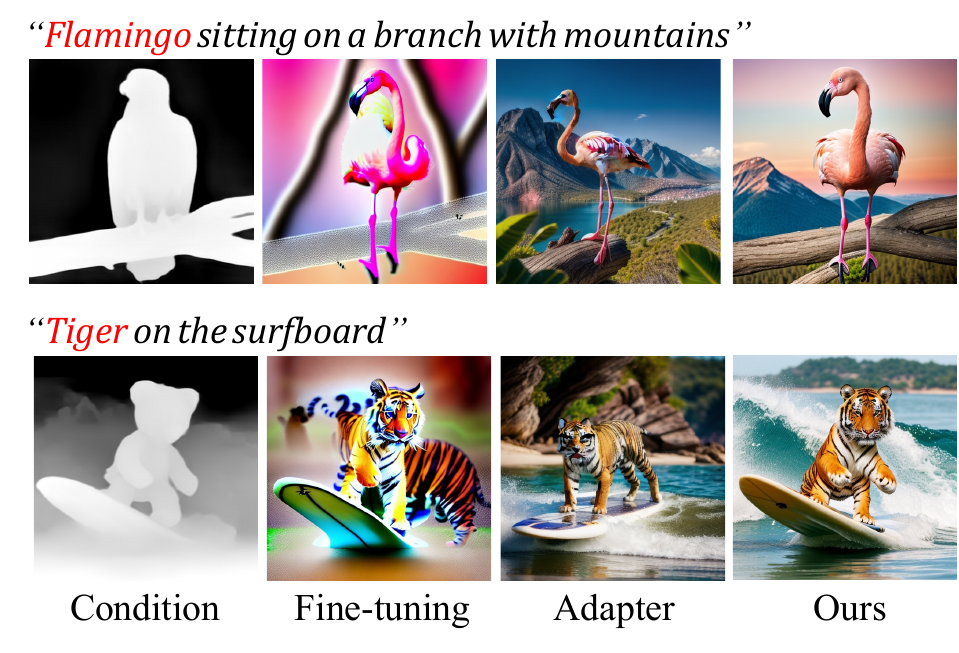}
    \vspace{-6mm}
    \caption{\scriptsize{Visual comparison for different implementation methods based on our dataset.}}
    \label{fig:SmartControl_model}
 \end{minipage}
 \vspace{-4mm}
\end{figure}

\noindent\textbf{Effect of Training Dataset Sizes}.
Even with a limited set of 0.5k images, our training process remains stable, and as the dataset size increases, the realism of the generated images improves (as shown in \cref{fig:SmartControl_datanum} and \cref{tab:SmartControl_datanum}). 
Obviously, we choose 2k images for our training dataset under the depth condition.
Although the dataset consists of only 2k images, we achieve commendable results across the open domain.
More analysis and visual results of generalization capability are provided in the supplementary material.

\captionsetup{font={scriptsize}, justification=raggedright}

\begin{figure}[t!]
  \centering
  \begin{minipage}[t]{0.49\textwidth}
    \centering
    \begin{overpic}[percent,width=\textwidth]{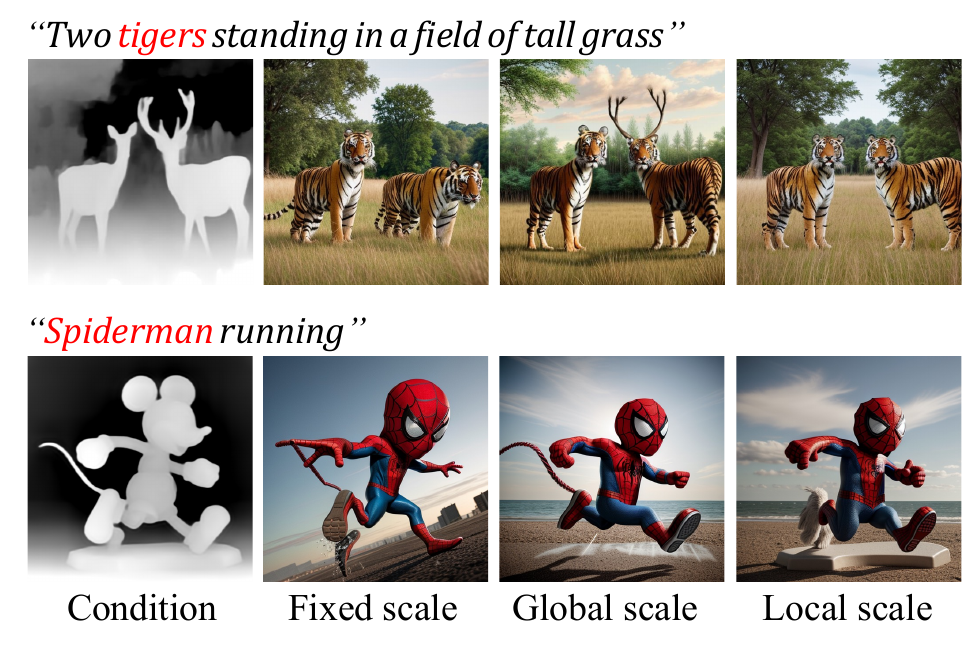}
    \end{overpic}
    \vspace{-6mm}
    \caption{\scriptsize{Visual comparison for the different granularity of control scale.}}
    \label{fig:SmartControl_c}
  \end{minipage}
  \hfill
  \centering
  \begin{minipage}[t]{0.49\textwidth}
    \centering
    \includegraphics[width=\textwidth]{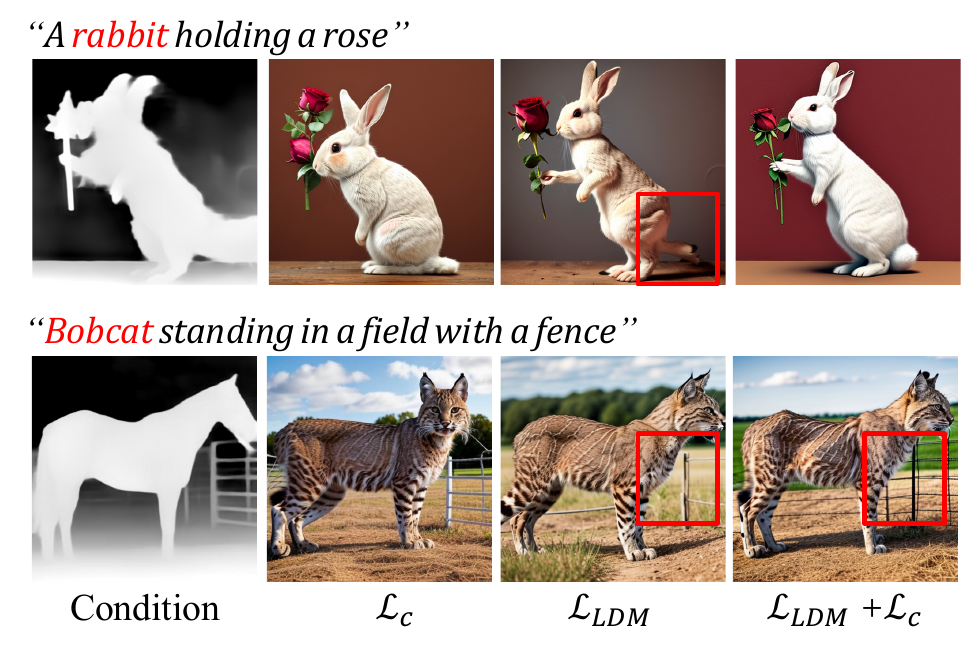} 
    \vspace{-6mm}
    \caption{\scriptsize{Impact of $\mathcal{L}_\mathrm{LDM}$ and $\mathcal{L}_\mathrm{c}$ contribute to the overall performance.} }
    \label{fig:SmartControl_loss}
  \end{minipage}  
\end{figure}

\begin{table}[t!]
    \begin{minipage}{0.49\textwidth}
        \centering
         \caption{\scriptsize{Ablation study on the sizes of training dataset under the depth condition.}}
         \vspace{-2mm}
        \scalebox{0.85}{
            \begin{tabular}{p{1.9cm}<{\centering} | p{1.2cm}<{\centering} | p{1.2cm}<{\centering} | p{1.2cm}<{\centering} }
            \hline
            Datasets & CLIP$\uparrow$  & CLASS$\uparrow$ &Sim$\downarrow$  \\
            \hline
            N=500 & 0.273 & 0.740 & 0.134 \\
            N=1000 & \textbf{0.274} & 0.724 & 0.130\\
            N=2000 & \textbf{0.274} & \textbf{0.742} & \textbf{0.128} \\
            \hline
            \end{tabular}
        }
        \label{tab:SmartControl_datanum}
    \end{minipage}%
    \begin{minipage}{0.49\textwidth}
        \centering
        \caption{\scriptsize{Effect of the proposed control scale predictor for rough conditions.}}
        \vspace{-2mm}
        \scalebox{0.85}{
            \begin{tabular}{p{1.9cm}<{\centering} | p{1.3cm}<{\centering} | p{1.3cm}<{\centering} | p{1.3cm}<{\centering} }
            \hline
            Method & CLIP$\uparrow$  & CLASS$\uparrow$ &Sim$\downarrow$  \\
            \hline
            Fine-tuning & 0.248&0.307&0.137 \\
            Adapter  & 0.273& 0.731&0.192  \\
            Ours &  \textbf{0.274} & \textbf{0.742} & \textbf{0.128}\\
            \hline
            \end{tabular}  
        }
        \label{tab:SmartControl_method_finetuning}
    \end{minipage}
    \vspace{-2mm}
\end{table}

\begin{table}[t!]
    \begin{minipage}{0.49\textwidth}
        \centering
        \caption{\scriptsize{Effect of local control scale.}}
        \vspace{-2mm}
        \scalebox{0.85}{
            \begin{tabular}{c|c|c|c}
            \hline
            Method & CLIP$\uparrow$  & CLASS$\uparrow$ &Sim$\downarrow$  \\
            \hline
            Fixed Scale $\alpha_{fix}$ & 0.268 & 0.710 & 0.136\\
            Global Scale $\alpha_{glob}$ &0.272 & 0.741& 0.122\\
            Local Scale $\bm{\alpha}$& \textbf{0.274} & \textbf{0.742} & \textbf{0.128}\\
            \hline
            \end{tabular}
        }
        \label{tab:local_scale}
    \end{minipage}%
    \begin{minipage}{0.49\textwidth}
        \centering
        \caption{\scriptsize{Ablation of the network architecture for the control scale predictor.}}
        \vspace{-2mm}
        \scalebox{0.85}{
            \begin{tabular}{c|c|c|c|c}
            \hline
            Method & CLIP$\uparrow$  & CLASS$\uparrow$ &Sim$\downarrow$  & Time$\downarrow$\\
            \hline
            Cross Atten & 0.272& 0.734&\textbf{0.126}& 7.69 \\
            Conv(Ours) & \textbf{0.274} & \textbf{0.742} & 0.128& \textbf{7.36} \\
            \hline
            \end{tabular}  
        }
        \label{tab:network_architecture}
    \end{minipage}
    \vspace{-2mm}
\end{table}

\vspace{0.5mm}
\noindent\textbf{Effect of Control Scale Predictor}.
As illustrated in \cref{sec:SmartControl_predictor}, we apply a control scale predictor to adaptively adjust the control intensity based on various conditions, and text prompts.
In this subsection, we make detailed experiments to assess the effects of the control scale predictor, \eg, the fine-tuning scheme, the granularity of the control scale, and the network structure.

\textit{Fine tuning scheme.}
In order to assess the effect on the control scale predictor, we experiment on several commonly used fine-tuning schemes, \eg, fine-tuning the ControlNet branch, an adapter, and a control scale predictor.
Fine-tuning the ControlNet may suffer from the degradation of generation capability.
This is evident from a performance drop in CLIP Score and the poor quality of the generated images.
On the other hand, training an adapter on our dataset may lead to overfitting, resulting in decreased structural alignment during testing as shown in \cref{fig:SmartControl_model}. 
As shown in \cref{tab:SmartControl_method_finetuning}, although CLIP Scores are comparable, our method achieves a 33.3\% improvement on the Self-similarity metric over the adapter, which demonstrates the effectiveness of the control scale predictor.

\textit{Granularity of control scale}.
In this subsection, we make detailed experiments to assess the effect of different granularity of control scale, \eg, fixed scale $a_\mathit{fix}$, global scale $a_\mathit{glob}$ and local scale map $\bm{\alpha}$.
Using a fixed scale across the entire evaluation dataset leads to a decrease in performance and the generation of lower-quality images.
We trained our model to predict the global scale based on our dataset.
However, the global scale is insufficient to handle situations where the required control scale varies within a single image.
For example, in the second line in \cref{fig:SmartControl_c}, the tail is not thoroughly removed, and the base part is not preserved.
In contrast, our method is designed to predict the local control scale, which effectively addresses these issues. 
As shown in \cref{tab:local_scale}, the performance is promoted with local control scale, which also demonstrates the effectiveness of the local control scale.

\textit{Network architecture}.
 We experiment with commonly used network architectures, \eg, convolution, and cross-attention.
The experimental results in \cref{tab:network_architecture} revealed that all of them yield better performance. 
We selected the convolution to implement the control scale predictor in this paper as it is relatively faster.

\vspace{0.5mm}
\noindent\textbf{Effects of Loss}.
We investigate the impact of $\mathcal{L}_\mathrm{LDM}$ and  $\mathcal{L}_\mathrm{c}$ in \cref{fig:SmartControl_loss}.
The model trained solely with $\mathcal{L}_\mathrm{LDM}$ exhibits relatively poor performance and lacks accuracy in constraining the layout.
Training without $\mathcal{L}_\mathrm{c}$ leads to the reduction of control even in non-conflicting areas, such as the background in the second row. 
Additionally, it results in residual artifacts such as the tail of the rabbit.


\section{Conclusion}

 In this paper, we introduce SmartControl, a flexible controllable image generation under rough visual conditions. 
Unlike existing approaches, SmartControl adaptively handles situations where there are disagreements between visual conditions and text prompts.
We introduce the control scale predictor, capable of identifying conflict regions between visual conditions and prompts and predicting local adaptive control scale based on the degree of conflict.
For training and evaluation, we construct a dataset with unaligned text prompts and visual conditions. 
Extensive experiments demonstrate that our SmartControl achieves better performance against state-of-the-art methods under rough visual conditions.

\clearpage  

%
%
\bibliographystyle{splncs04}
\bibliography{egbib}
\end{document}